\documentclass[10pt,twocolumn,letterpaper]{article}

\usepackage{iccv}
\usepackage{times}
\usepackage{epsfig}
\usepackage{graphicx}
\usepackage{amsmath}
\usepackage{amssymb}
\usepackage[accsupp]{axessibility}  

\usepackage[breaklinks=true,bookmarks=false]{hyperref}

\iccvfinalcopy 

\ificcvfinal\fi

\begin{document}

\title{Geometry-guided Feature Learning and Fusion for Indoor Scene Reconstruction}

\author{Ruihong Yin$^{1}$, \hspace{2mm} Sezer Karaoglu$^{1,2}$, \hspace{2mm} Theo Gevers$^{1,2}$\\
$^{1}$University of Amsterdam, Amsterdam, The Netherlands \\ $^{2}$3DUniversum, Amsterdam, The Netherlands\\
{\tt\small r.yin@uva.nl, s.karaoglu@3duniversum.com, Th.Gevers@uva.nl}
}

\maketitle
\ificcvfinal\thispagestyle{empty}\fi

\begin{abstract}
   In addition to color and textural information, geometry provides important cues for 3D scene reconstruction. However, current reconstruction methods only include geometry at the feature level
   thus not fully exploiting the geometric information. 
   
   In contrast, this paper proposes a novel geometry integration mechanism for 3D scene reconstruction. Our approach incorporates 3D geometry at three levels, i.e. feature learning, feature fusion, and network supervision. First, geometry-guided feature learning encodes geometric priors to contain view-dependent information. Second, a geometry-guided adaptive feature fusion is introduced which utilizes the geometric priors as a guidance to adaptively generate weights for multiple views. Third, at the supervision level, taking the consistency between 2D and 3D normals into account, a consistent 3D normal loss is designed to add local constraints.
   
   Large-scale experiments are conducted on the ScanNet dataset, showing that volumetric methods with our geometry integration mechanism outperform state-of-the-art methods quantitatively as well as qualitatively. Volumetric methods with ours also show good generalization on the 7-Scenes and TUM RGB-D datasets.   
\end{abstract}

\section{Introduction}

3D scene reconstruction is an important topic in 3D computer vision, with many applications such as mixed/augmented reality, autonomous navigation, and robotics. It is also considered one of the fundamental tasks in 3D scene understanding including 3D segmentation \cite{robert2022learning, tang2022contrastive, vu2022softgroup} and object detection \cite{wang2022cagroup3d, wang2022multimodal}. Although nowadays cameras equipped with depth sensors (\eg Lidar and Kinect) can reconstruct scenes using perspective projection and depth fusion \cite{newcombe2011kinectfusion}, these $RGB\text{-}D$ cameras are still expensive, and not yet widely used in consumer cameras. Therefore, they are limited in their applicability. In contrast, scene reconstruction from $RGB$ images (multi-view or video) is much more accessible. 

\begin{figure}[t]
  \centering

  \includegraphics[width=1.0\linewidth]{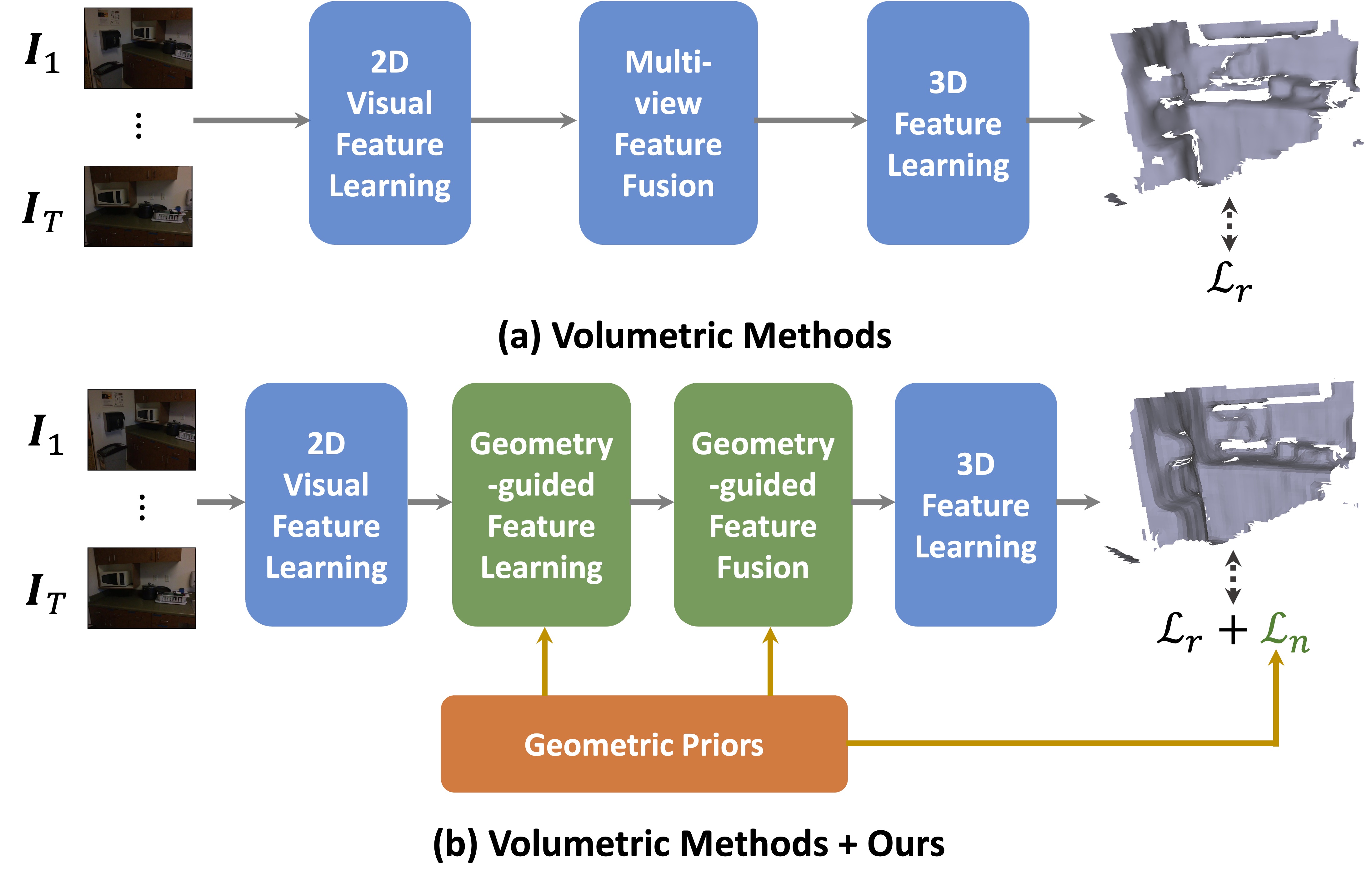}
  \caption{{\textbf{Pipeline of existing volumetric methods compared to our proposed geometry-guided feature learning and fusion for 3D scene reconstruction.}} Our approach (green parts) integrates view-dependent and local geometry into (1) feature learning, (2) multi-view feature fusion, and (3) network supervision.}
  \label{fig: motivation}
\end{figure}

A standard approach to 3D scene reconstruction is to compute the Truncated Signed Distance Function (TSDF) volume and then apply the marching cubes algorithm \cite{lorensen1987marching} to capture the surface. To generate the TSDF volume, traditional reconstruction methods \cite{im2019dpsnet, yao2018mvsnet, yao2019recurrent, sayed2022simplerecon, wang2018mvdepthnet, yi2020pyramid} first generate depth maps for each $RGB$ image and then apply depth fusion \cite{curless1996volumetric}. Due to pixel-level prediction, depth-based methods can generate dense 3D points but may suffer from scale ambiguity and depth inconsistency between overlapping regions in different views. {Recently, volumetric (direct) methods \cite{murez2020atlas, liu2019neural, bozic2021transformerfusion, stier2021vortx} are proposed to predict the TSDF directly, without reliance on depth estimation. 3D scenes are modeled using volumetric methods that employ 3D CNNs, allowing for the filling of unobserved gaps and resulting in enhanced predictions.} However, both depth-based and volumetric methods still capture texture and color features based on $RGB$ information. 

For multi-view tasks, geometric information (\eg surface normal and viewing direction) provides rich view-dependent cues of 3D scenes. For example, the best viewing direction is perpendicular to the viewing position. This viewpoint (or one close to it) is preferred over other views. Also, voxels derived from the same plane should have similar surface normals. Hence, extracting important cues from these geometries can be beneficial for feature learning and scene representation. In addition, multi-view feature fusion plays a vital role in volumetric reconstruction methods. Due to changing imaging conditions (\eg illumination, camera orientation, and occlusion), instead of simply averaging views, some views may be preferred over others in terms of their positioning (\ie more useful geometry information). Furthermore, volumetric methods usually supervise the predicted TSDF in a voxel-to-voxel manner ignoring local information, and hence may deviate from the actual surfaces. 

To address the aforementioned issues, in this paper, a geometry integration mechanism is proposed for 3D scene reconstruction. To this end, geometric information is exploited by our method at three different stages (see Figure \ref{fig: motivation}{b}): (1) feature learning, (2) feature fusion, and (3) network supervision. Firstly, to exploit discriminative information for 3D reconstruction, a geometry-guided feature learning (G2FL) is introduced to encode and integrate geometric priors (\eg surface normal, projected depth, and viewing direction) into the multi-view features. Transformers and multi-layer perceptron (MLP) are utilized to exploit the geometric information. Secondly, during multi-view feature fusion, the occluded views and views away from others may be assigned different attention levels. Therefore, the occlusion prior and relative pose distance are adopted to construct the multi-view attention function, forming a geometry-guided adaptive feature fusion (G2AFF). Thirdly, at the supervision level, the 3D surface normal is calculated from the TSDF, which at the same time maintains local information. To enhance the local constraints and improve the reconstruction quality, a consistent 3D normal loss (C3NL) is proposed, considering the consistency between 2D and 3D normal, discarding boundaries and thin objects. 

Our main contributions are summarized as follows:
\begin{itemize}
    \item A novel geometry integration mechanism is proposed for 3D scene reconstruction, encoding geometric priors at three levels, \ie feature learning, feature fusion, and network supervision.
    \item A geometry-guided feature learning scheme encodes 3D geometry into multi-view features. A geometry-guided adaptive feature fusion method uses geometric priors as a guidance to learn a multi-view weight function adaptively.
    \item The consistency between 2D and 3D normal is exploited. A consistent 3D normal loss is introduced to constrain local planar regions in the prediction.
    \item Volumetric methods enhanced with our method show state-of-the-art performance on the ScanNet dataset and demonstrates convincing generalization on the 7-Scenes and TUM RGB-D datasets.
\end{itemize}

\section{Related work}

\subsection{3D scene reconstruction}
\noindent \textbf{Depth-based reconstruction.} Depth-based methods typically follow a similar approach, \ie first building a plane sweep cost volume \cite{collins1996space, gallup2007real} at the image or feature level, and then using convolutional layers to extract and fuse features from neighbouring views, finally predicting the depth maps. Cost volume aims to capture information from source images, as complementary features for the reference image. Different cost metrics are used, \eg concatenation, dot product, and per-channel variance. For example, MVSNet \cite{yao2018mvsnet} proposes a variance-based cost in each channel. In DPSNet \cite{im2019dpsnet}, the cost is calculated by concatenating reference features and the warped features. MVDepthNet \cite{wang2018mvdepthnet} and GP-MVS \cite{hou2019multi} adopt absolute differences between input images to measure the similarity of different views, while Neural RGBD \cite{liu2019neural} uses the same metric at the feature level. DeepVideoMVS \cite{duzceker2021deepvideomvs} and SimpleRecon \cite{sayed2022simplerecon} compute the dot product between reference and warped features. 

\noindent \textbf{Volumetric (direct) reconstruction.} Atlas \cite{murez2020atlas} is the first work to regress the TSDF directly, without the depth map as an intermediate product. Compared to depth-based methods, Atlas learns to fill in unobserved regions. Based on Atlas, NeuralRecon \cite{sun2021neuralrecon} designs a learning-based TSDF fusion to transfer features from previous to current fragments. TransformerFusion \cite{bozic2021transformerfusion} proposes a learned multi-view fusion module using a Transformer and predicts the occupancy similar to \cite{peng2020convolutional}. VoRTX \cite{stier2021vortx} adopts a Transformer to extract features and proposes an occlusion-aware fusion module.

\begin{figure*}[t]
  \centering
  \includegraphics[width=1.0\linewidth]{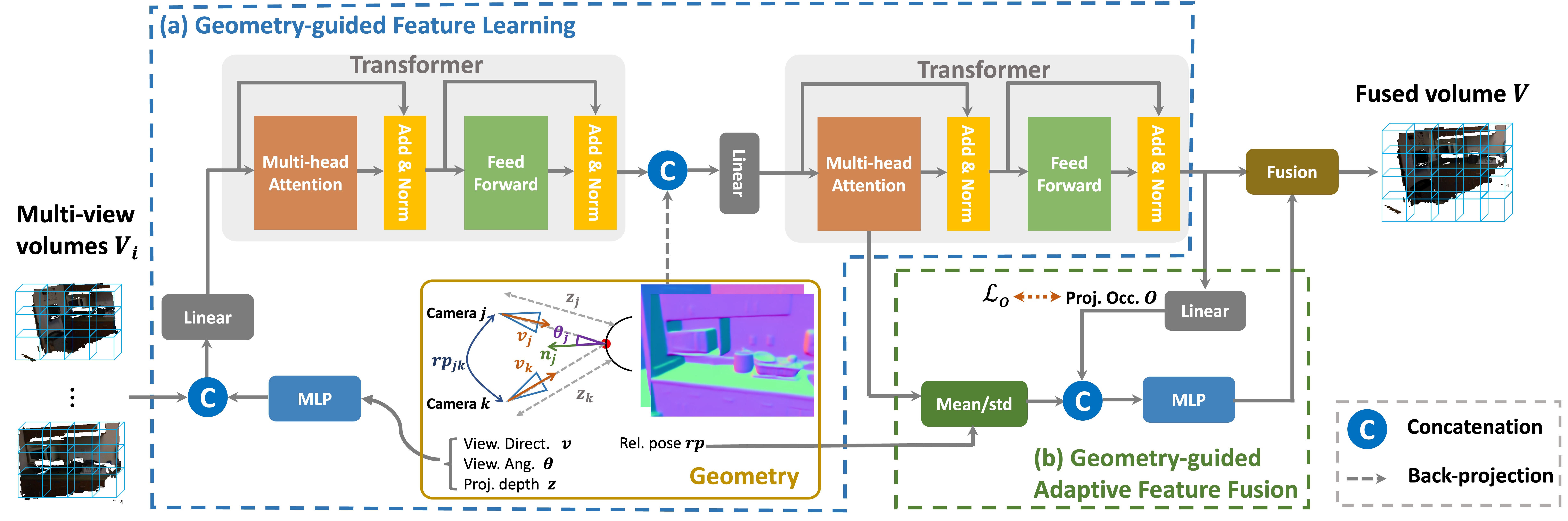}
  \caption{\textbf{Details of our proposed geometry-guided feature learning and geometry-guided adaptive feature fusion.} (a) Geometry-guided feature learning: After 2D visual feature learning, view-dependent geometric priors (\eg surface normal and viewing direction) are encoded and fused into the visual features of the multi-view volume using a MLP, linear layers, and Transformers. (b) Geometry-guided adaptive feature fusion: Fusion weighting is adaptively learned by a MLP with the guidance of features, relative pose distances, and occlusion priors.}
  \label{fig: framework}
\end{figure*}

\subsection{Geometric priors in 3D scene reconstruction}
A number of methods use geometric information for 3D scene reconstruction. For example, GP-MVS \cite{liu2019neural} applies a relative pose distance to the Gaussian kernel \cite{seeger2004gaussian}, which guides the learning in latent space. NeuralRecon \cite{sun2021neuralrecon} concatenates the projected depth to 3D features after multi-view fusion. TransformerFusion \cite{bozic2021transformerfusion} integrates pixel validity, viewing direction, and projected depth into the features. Viewing direction and occlusion prior are exploited in VoRTX \cite{stier2021vortx}. SimpleRecon \cite{sayed2022simplerecon} introduces the use of geometric metadata for scene reconstruction, \eg ray angle and depth validity mask. However, they only use a limited number of geometric priors and exploit 3D geometry at the feature level. In contrast, our method proposes to exploit geometric priors at different stages of the 3D scene reconstruction pipeline.

\subsection{Multi-view feature fusion}
The standard way of fusing multi-view features, \ie computing the average, considers each view in the same way. In contrast, attention-based fusion gives attention according to the information in each view. For instance, the AttnSets module \cite{yang2020robust} is proposed to aggregate multi-view features. DeepVideoMVS \cite{duzceker2021deepvideomvs} includes a ConvLSTM \cite{shi2015convolutional} to integrate past information into the current view. In particular, the use of Transformers \cite{vaswani2017attention, chu2021twins, wang2021pyramid, zhang2021vidtr} shows their effectiveness in feature awareness. Also, other methods are proposed, designing their fusion module based on Transformers. For example, TransformerFusion \cite{bozic2021transformerfusion} adopts Transformers to learn weights for each view and to select views during inference. VoRTX \cite{stier2021vortx} constructs an occlusion-aware fusion using Transformers. In contrast to existing methods, in this paper, an adaptive feature fusion is proposed to model the attention by the guidance of multi-view features and geometries.

\begin{figure*}[t]
  \centering
  \includegraphics[width=1.0\linewidth]{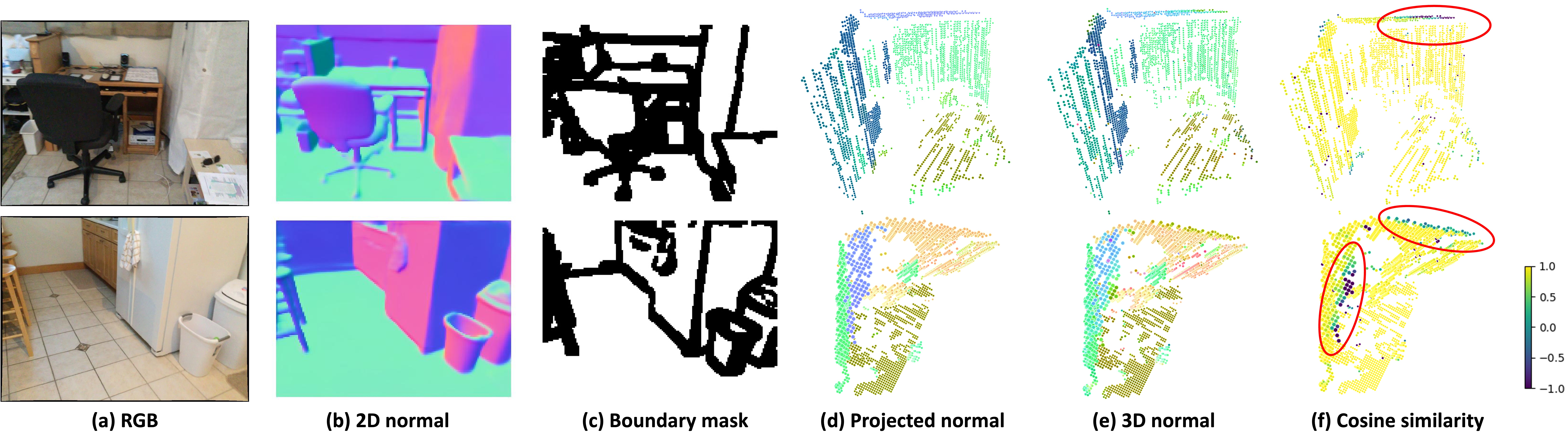}
  \caption{\textbf{Boundary and consistency analysis of our proposed  3D normal loss.} (a) $RGB$ images. (b) 2D surface normals predicted by a pre-trained normal network \cite{bae2021estimating}. (c) 2D boundary masks. White regions are planes, which are retained for normal loss computation. (d) Projected normal $\widetilde{\textbf{N}}$ is the 3D normal back-projected from the 2D normal. (e) 3D normal $\textbf{N}$ is generated from the ground truth of the TSDF, showing noise near the boundaries. (f) Cosine similarity between (d) and (e). Blue points in the red circle mean that angles between (d) and (e) are greater than $90^{\circ}$. }
  \label{fig: normal}
\end{figure*}
\section{Method} 
$T$ images $\textbf{I}_i\in \mathbb{R}^{3 \times H \times W}$ with camera intrinsics $\textbf{K}_i\in \mathbb{R}^{3 \times 3}$ and camera pose $\textbf{P}_i=(\textbf{R}_i, \textbf{t}_i) \in \mathbb{R}^{4 \times 4}$ are taken as an input, where $i$ is the view index. As shown in Figure \ref{fig: motivation}{a}, volumetric (direct) methods generally consist of three core components, \ie 2D visual feature learning, multi-view feature fusion, and 3D feature learning. 2D visual feature learning exploits 2D convolutional neural networks (CNNs) \cite{he2016deep, tan2019mnasnet} to extract 2D features, after which they are back-projected to a 3D space. Next, multi-view feature fusion combines these features into one volume. Finally, 3D feature learning adopts a 3D CNN \cite{tang2020searching} to regress the TSDF value. Our geometry integration mechanism aims to combine 3D geometry into general volumetric methods, see Figure \ref{fig: motivation}{b}. The key differences are: \textbf{(1)} After 2D visual feature learning, geometry-guided feature learning incorporates view-dependent geometric information (\eg surface normal and viewing direction) into the 3D volume, which is processed by Transformers and a MLP to exploit useful cues. Details are given in Section \ref{sec: feat learn}. \textbf{(2)} In the multi-view fusion stage, a geometry-guided adaptive feature fusion is proposed. Features, occlusion approximation, and relative pose distances are used to guide view-attention learning. The multi-view volumes are integrated into one volume by learned weights. Section \ref{sec: feat fusion} outlines this approach. \textbf{(3)} In the loss function, the 3D normal calculated from the TSDF contains local information of the TSDF. To encourage the network to generate consistent scenes, a 3D normal loss is added to the output. The normal loss keeps consistency between 2D and 3D normals and ignores boundaries and thin parts. The normal loss is only computed during training. Section \ref{sec: supervision} provides more details about this stage.
\subsection{Geometry-guided feature learning}
\label{sec: feat learn}
Planar structures are common in indoor scenes, \eg walls and tables. Hence, surface normals provide vital information to determine the relationship between planes. Due to back-projection, voxels along the camera ray correspond to the same 2D features. Thus, depth can add discriminative cues, \eg voxels close to the camera provide more details, while distant voxels contain richer contextual information. Furthermore, the viewing direction corresponds to the orientation of voxels in a camera coordinate frame, which is also related to the amount of camera distortion. Therefore, projected surface normal (back-projected from 2D normal), viewing angle, projected depth (calculated from the voxel coordinate by perspective projection), and viewing direction are all informative for indoor scenes. (More details about geometry calculation can be found in the supplementary material). In Figure \ref{fig: framework}{a}, these geometric priors are explicitly integrated in the feature learning process.

In our approach, the use of normal and other geometry information is exploited by two separate modules, \ie $\mathcal{T}_1$ and $\mathcal{T}_2$. In $\mathcal{T}_1$, to provide high-frequency information, the viewing direction $\textbf{v}_i\in \mathbb{R}^{3 \times N_v}$ ($N_v$ is the number of voxels) and projected depth $\textbf{z}_i\in \mathbb{R}^{1 \times N_v}$ are encoded similar to NeRF \cite{mildenhall2020nerf}. Then, the encoded priors $\gamma(\textbf{v}_i)\in \mathbb{R}^{(6L) \times N_v}$ and $\gamma(\textbf{z}_i)\in \mathbb{R}^{(2L) \times N_v}$ are concatenated using the original viewing direction $\textbf{v}_i$, projected depth $\textbf{z}_i$, and viewing angle ${\boldsymbol{\theta}}_i$. After this, they are processed by a MLP. Next, the processed geometry is concatenated to the 3D volume ${\textbf{V}}_i\in \mathbb{R}^{C_v \times N_v}$. Then, a linear layer is applied to reduce the channel dimension generating $\textbf{g}_i\in \mathbb{R}^{C_v \times N_v}$. Finally, $\textbf{g}_i$ is used as input of a Transformer to further combine visual and geometric features.
\begin{equation}
\begin{aligned}
\mathcal{T}_1: \textbf{g}_i &= \text{Linear} (\lbrack \text{MLP} (\lbrack \gamma(\textbf{v}_i), \gamma(\textbf{z}_i), \textbf{v}_i, \textbf{z}_i, \boldsymbol{\theta}_i \rbrack), \textbf{V}_i \rbrack) \\
\boldsymbol{\varphi}_i &= \text{Transformer} (\textbf{g}_i) \\
\end{aligned}
\end{equation}
where $\lbrack \cdot \rbrack$ denotes channel-wise concatenation.

In $\mathcal{T}_2$, the projected surface normal $\textbf{n}_i\in \mathbb{R}^{3 \times N_v}$ is concatenated with feature $\boldsymbol{\varphi}_i\in \mathbb{R}^{C_v \times N_v}$. Then, a linear layer is applied to reduce the dimension and to combine normal with previous features. Another Transformer is adopted to integrate the geometry and visual features. In particular, Transformers in $\mathcal{T}_1$ and $\mathcal{T}_2$ are applied in a temporal manner, also exploiting information between multiple views.
\begin{equation}
\begin{aligned}
\mathcal{T}_2: \boldsymbol{\phi}_i = \text{Transformer} (\text{FCN} ([\textbf{n}_i, \boldsymbol{\varphi}_i])) 
\end{aligned}
\end{equation}

\subsection{Geometry-guided adaptive feature fusion}
\label{sec: feat fusion}
In back-projection, the occluded 3D voxel may be mapped to an irrelevant pixel (\ie 2D features) adding noise to the feature fusion module. Moreover, although voxels between the camera and the surface are given, empty space regions are useless for the reconstruction task. As a result, projective occupancy is adopted as an approximation of occlusion, which also allows features to include relative depth information. After $\mathcal{T}_2$, the projective occupancy probabilities are predicted by a linear layer and sigmoid function as follows, 
\begin{equation}
    \textbf{O}_i = \text{Sigmoid} (\text{Linear}(\boldsymbol{\phi}_i)) 
\end{equation}
Binary cross-entropy loss is applied on $\textbf{O}_i\in \mathbb{R}^{1 \times N_v}$ as an occupancy loss $\mathcal{L}_o$ to supervise the prediction.

The Transformer in $\mathcal{T}_2$ computes an attention matrix $\textbf{A}\in \mathbb{R}^{T \times T}$ for each voxel, in which each row $\textbf{A}_i$ represents the relationship between the \emph{i}th and other views. If the weights in the same row are similar (the row-wise standard deviation is close to 0), this means that the view has the same features as the other views. Conversely, if the row-wise weights are different from each other (the row-wise standard deviation is large), this implies that the view may carry important features. Additionally, if the camera location of the view is distant from the others, the voxel in this view may contain distinctive features, or the voxel may be occluded. Therefore, the occlusion prior, relative pose distance, and the row-wise statistics information (\ie mean and standard deviation) of the attention matrix can provide vital information for determining the best views. To this end, a geometry-guided adaptive feature fusion is designed to generate weights for multiple views. The mean and standard deviation for the attention matrix and relative pose distance are first computed. Then, they are concatenated based on projective occupancy probabilities. A MLP is applied to filter the noise and adaptively learn the weight from the visual features and geometric priors. Finally, a softmax is used to generate the weight:
\begin{equation}
\begin{aligned}
    &\textbf{w} = \text{Softmax} (\text{MLP}([\boldsymbol{\mu} ^A, \boldsymbol{\sigma}^A, \boldsymbol{\mu}^{rp}, \boldsymbol{\sigma}^{rp}, \textbf{O}])) \\
    &\textbf{V} = \sum_{i} \textbf{w}_{i}\textbf{V}_{i}
\end{aligned}
\end{equation}
where $\boldsymbol{\mu}^A\in \mathbb{R}^{T \times N_v}$ and $\boldsymbol{\sigma}^A\in \mathbb{R}^{T \times N_v}$ represent the row-wise mean and standard deviation of the attention matrix. $\boldsymbol{\mu}^{rp}\in \mathbb{R}^{(3T) \times N_v}$ and $\boldsymbol{\sigma}^{rp}\in \mathbb{R}^{(3T) \times N_v}$ denote the mean and standard deviation of the relative pose distance.
\subsection{Consistent 3D normal loss}
\label{sec: supervision}

When the voxel $\textbf{p}\in\mathbb{R}^{3 }$ is on or close to the surface (TSDF $S$ is 0), the numerical derivative of the TSDF is the 3D surface normal $\textbf{N}(\textbf{p})\in\mathbb{R}^{3 \times 1}$ of the voxel \cite{newcombe2011kinectfusion}:
\begin{equation}
    \textbf{N}(\textbf{p})=\nu [\nabla {S(\textbf{p})}], \nabla {S(\textbf{p})}=[\frac{\partial {S}}{\partial x}, \frac{\partial {S}}{\partial y}, \frac{\partial {S}}{\partial z}]^T. \\
\label{eq: 3D normal}
\end{equation}
where $\nu [\textbf{x}] = \textbf{x} / \Arrowvert \textbf{x} \Arrowvert _2$. In our approach, the numerical derivative is implemented by a derivative operator \cite{prewitt1970object}. In this way, the ground truth of the 3D normal $\textbf{N}_{gt}$ is generated from the ground truth of TSDF by Eq. \ref{eq: 3D normal}, while the prediction of the 3D normal $\textbf{N}_{pred}$ corresponds to the predicted TSDF.

Because normals near the boundaries or thin/small objects are usually inaccurate, a boundary mask $\textbf{M}_{2d}\in \mathbb{R}^{1 \times H_{f} \times W_{f}}$ is introduced to filter out these parts and to ensure that the normal loss is only calculated for planar regions. $\textbf{M}_{2d}$, shown in Figure \ref{fig: normal}{c}, is calculated as follows: a 2D edge detector \cite{irwin1968isotropic} is used to compute gradients at 2D normals. Then pixels with gradient values greater than a threshold are regarded as boundary pixels. Finally, an 8 neighbor of boundary pixels is also regarded as boundary pixels (implemented by a max-pooling layer with kernel size 3 and stride 1). $\textbf{M}_{2d}$ is back-projected to 3D space, forming the 3D boundary mask $\textbf{M}_{3d}\in \mathbb{R}^{1 \times N_v}$. After masking the voxels by $\textbf{M}_{3d}$, other noise sources in the computation of 3D normal may exist, as shown in Figure \ref{fig: normal}{e} and \ref{fig: normal}{f}. Hence, this paper introduces a consistency between projected normal $\widetilde{\textbf{N}}(\textbf{p})$ and 3D normal $\textbf{N}(\textbf{p})$ to suppress noise. The projected surface normal $\textbf{n}_i$, see Section \ref{sec: feat learn}, is in camera coordinates while $\textbf{N}(\textbf{p})$ is in world coordinates. Therefore, $\textbf{n}_i$ is transformed to world coordinates by $\widetilde{\textbf{N}}_i(\textbf{p})=\textbf{R}_i \textbf{n}_i$ and then averaged between views by $\widetilde{\textbf{N}}(\textbf{p})= \frac{1}{T} \sum_{i=1}^T \widetilde{\textbf{N}}_i(\textbf{p})$. $s_{2d3d}$ in Eq. \ref{eq: consistency} computes the cosine similarity between $\textbf{N}(\textbf{p})$ and $\widetilde{\textbf{N}}(\textbf{p})$ to measure consistency.
\begin{equation}
    s_{2d3d}(\textbf{p}) = \frac{\textbf{N} \cdot \widetilde{\textbf{N}}}{\Arrowvert \textbf{N} \Arrowvert _2 \Arrowvert \widetilde{\textbf{N}} \Arrowvert _2}
    \label{eq: consistency}
\end{equation}
The indicator function $[s_{2d3d}(\textbf{p})>0]$ is applied to generate the consistency measure, \ie if the 3D surface normal $\textbf{N}(\textbf{p})$ is similar to the projected 3D surface normal $\widetilde{\textbf{N}}(\textbf{p})$, the normal loss is computed. Otherwise, the 3D normal is considered  as noise. The weight $W_{2d3d}(\textbf{p})$ is given by 
\begin{equation}
     W_{2d3d}(\textbf{p})= [s_{2d3d}(\textbf{p})> 0] \equiv
    \begin{cases}
        1,& s_{2d3d}(\textbf{p})> 0 \\
        0, & s_{2d3d}(\textbf{p})\le 0
    \end{cases}
\label{eq: consist. weight}
\end{equation}

By excluding boundary voxels and considering consistency between projected and 3D normals, the 3D normal loss is defined by:
\begin{equation}
     \mathcal{L}_n = 1-\frac{1}{N}\sum_{m=1}^N M_{3d}^m W_{2d3d}^m \frac{\textbf{N}_{gt}^{m} \cdot {\textbf{N}}_{pred}^m}{\Arrowvert \textbf{N}_{gt}^{m} \Arrowvert _2 \Arrowvert {\textbf{N}}_{pred}^m \Arrowvert _2} 
\end{equation}
where $N$ is the number of voxels used in the normal loss.
\section{Experiments} 
\begin{table}[t]
  \small
\setlength\tabcolsep{2pt}
  \centering
  \begin{tabular}{cccccc}
    \hline
    Method  & Comp$\downarrow$ & Acc$\downarrow$ &Recall$\uparrow$ &Prec$\uparrow$ &F-score$\uparrow$   \\
    \hline
    
    COLMAP \cite{schonberger2016pixelwise} &  0.069 &0.135 &0.634 &0.505 &0.558  \\
    MVDepthNet\cite{wang2018mvdepthnet} & \underline{0.040} &0.240 &\underline{0.831} &0.208 &0.329 \\
    GPMVS \cite{hou2019multi} & \textbf{0.031} &0.879 &\textbf{0.871} &0.188 &0.304   \\
    DPSNet \cite{im2019dpsnet} & 0.045 & 0.284 & 0.793 &0.223 &0.344    \\
    SimpleRecon \cite{sayed2022simplerecon} & 0.078 &0.065 &0.641 &0.581 &{0.608} \\
     Atlas \cite{murez2020atlas} & 0.084 &0.102 &0.598 &0.565 &0.578     \\
     TransformerFusion \cite{bozic2021transformerfusion} & 0.099 &0.078 &0.648 &0.547 &0.591 \\
    3DVNet \cite{rich20213dvnet} & 0.077 &0.221 &0.506 &0.545 &0.520 \\
    \hline
    NeuralRecon \cite{sun2021neuralrecon}& 0.138 & \underline{0.051} &0.478 & {0.683} &0.560 \\
    {NeuralRecon + ours} & 0.099 &\textbf{0.048} &0.545 & \textbf{0.722}&\underline{0.619} \\
    {VoRTX} \cite{stier2021vortx} & 0.108 &0.062 & 0.545&0.666 &0.598 \\
    {VoRTX + Ours} & 0.098 &{0.059} &0.585 &\underline{0.687} &\textbf{0.630} \\
    \hline
  \end{tabular}
  \caption{3D reconstruction mesh evaluation following Atlas \cite{murez2020atlas} for ScanNet. The best results are \textbf{bold}, and the second best ones are \underline{underlined}. }
  \label{tab:recon}
\end{table}

\begin{table}[t]
  \small
\setlength\tabcolsep{3pt}
  \centering
  \begin{tabular}{cccccc}
    \hline
    \multicolumn{6}{c}{\textbf{7-Scenes}} \\ \hline
    Method  & Comp$\downarrow$ & Acc$\downarrow$ &Prec$\uparrow$ &Recall$\uparrow$ &F-score$\uparrow$   \\
    \hline
    
    DeepV2D \cite{teed2018deepv2d} &  \underline{0.180} &0.518 &0.087 &0.175 &0.115  \\
    CNMNet\cite{long2020occlusion} & \textbf{0.150} &0.398 &{0.111} &0.246 &0.149 \\
    
    NeuralRecon \cite{sun2021neuralrecon} & {0.228} &\underline{0.100} &\underline{0.389} &0.227 &0.282   \\
    {NeuralRecon + ours}  & 0.289 & \textbf{0.086} & \textbf{0.476} &\underline{0.294} &\textbf{0.359}    \\
    VoRTX \cite{stier2021vortx} & {0.286} &{0.103} &{0.364} &{0.267} &{0.304}   \\
    {VoRTX + ours}  & 0.231& \underline{0.100} & {0.381} &\textbf{0.299} &\underline{0.332}    \\
    \hline
    \multicolumn{6}{c}{\textbf{TUM RGB-D}} \\ \hline
    Method  & Comp$\downarrow$ & Acc$\downarrow$ &Prec$\uparrow$ &Recall$\uparrow$ &F-score$\uparrow$   \\
    \hline
     Atlas \cite{murez2020atlas} & 2.344 &0.208 &0.360 &{0.089} &0.132     \\
     
     NeuralRecon \cite{sun2021neuralrecon}& 1.341 & \underline{0.092} &\textbf{0.564} &0.155 &{0.232} \\
    {NeuralRecon + ours} & \underline{0.851} &\textbf{0.087} &\underline{0.517} &{0.175} &{0.256} \\
    VoRTX \cite{stier2021vortx} & {0.911} &0.136 &0.434&\underline{0.203} & \underline{0.268} \\
    {VoRTX + ours}  & \textbf{0.722} & {0.128} & {0.445} &\textbf{0.217} &\textbf{0.284}    \\
    \hline
  \end{tabular}
  \caption{3D reconstruction mesh evaluation following Atlas \cite{murez2020atlas} on the 7-Scenes and TUM RGB-D datasets. }
  \label{tab:generalization}
  \end{table}
\subsection{Datasets and metrics}
Our method is evaluated on three challenging indoor $RGB\text{-}D$ datasets, \ie ScanNet(V2) \cite{dai2017scannet}, 7-Scenes \cite{shotton2013scene}, and TUM RGB-D \cite{sturm12iros} datasets. ScanNet consists of 807 unique indoor scenes, which is composed of 1613 scans (1201 for training, 312 for validation, and 100 for testing). Our method is trained on the training set of ScanNet. To validate the generalization, the method is also tested on 7-Scenes and TUM RGB-D datasets without fine-tuning. The ground-truth meshes for 7-Scenes and TUM RGB-D are produced by TSDF fusion with a voxel size of 4cm.

For quantitative comparison, 3D geometry metrics defined by \cite{murez2020atlas} are adopted to measure the quality of 3D reconstruction, including accuracy (acc), completeness (comp), precision (prec), recall, and F-score. F-score is considered the most reliable metric. The computation of each metric is detailed in the supplementary material.

\subsection{Implementation details} 
The online method NeuralRecon and the offline method VoRTX are chosen as our baselines. The number of heads in the Transformer is 2. The weights for occupancy loss, TSDF loss, projective occupancy loss, and 3D normal loss are $\{1.5, 1.0, 0.5, 0.1\}$. At the start of training, predicted TSDF may be inaccurate, causing a high 3D normal loss. Hence, the 3D normal loss is added after 5 epochs. The batch size per GPU is 4. Other settings (\eg view selection and voxel size) are similar to baselines. The network is trained on three NVIDIA RTX A6000 GPUs. The 2D surface normal is predicted by the pre-trained model in \cite{bae2021estimating}.
\subsection{Evaluation results}

\begin{figure*}[t]
  \centering
  \includegraphics[width=0.97\linewidth]{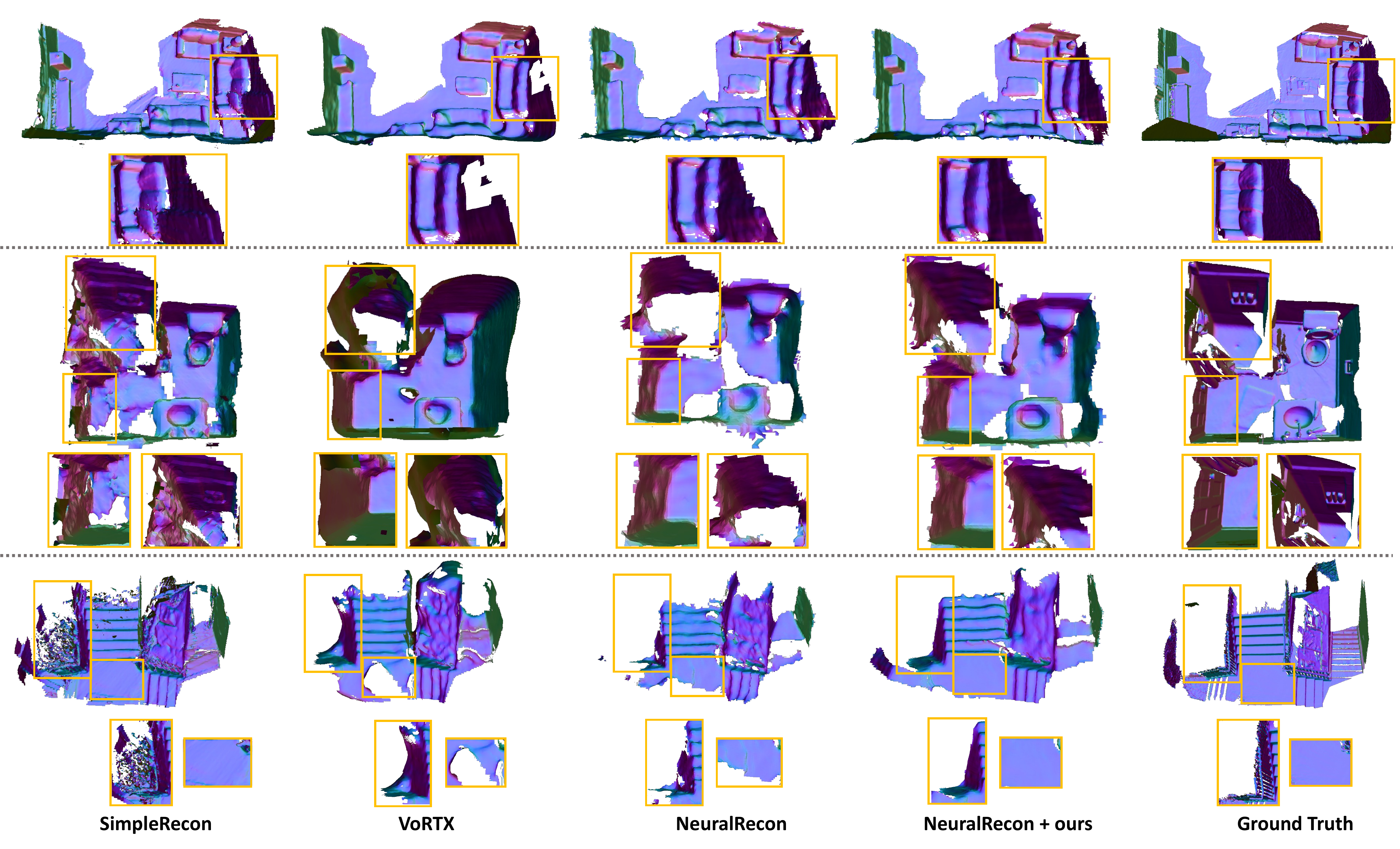}
  \caption{\textbf{Qualitative results on ScanNet.} Colors on the meshes are related to surface normals. Compared to other methods, NeuralRecon + ours is able to generate more regions, smoother planes, and more accurate geometry relationships.
  }
  \label{fig: visual}
\end{figure*}

\noindent \textbf{ScanNet \cite{dai2017scannet}.} Comparison between our version and other SOTA methods is shown in Table \ref{tab:recon}. When contrasted with NeuralRecon and VoRTX, both enhanced with our approach, NeuralRecon + our method and VoRTX + our method exhibit better performance across all 3D metrics. For example, F-score, precision, recall of NeuralRecon + ours  are 5.9\%, 3.9\%, 6.7\% higher than NeuralRecon. This is because our geometry integration mechanism adds more information to the voxels. VoRTX + ours and NeuralRecon + ours outperform SOTA methods in precision and F-score. In particular, compared to the depth-based method SimpleRecon, NeuralRecon + ours shows strong accuracy  (26.2\% decrease) and precision (14.1\% increase) performances. VoRTX + ours outperforms the volumetric method TransformerFusion by 14.0\% in precision, 24.4\% in accuracy, and 3.9\% in F-score. NeuralRecon + ours also outperforms VoRTX on almost all metrics. Qualitative results are presented in Figure \ref{fig: visual}. It is shown that NeuralRecon falls short in a number of regions (\eg floors). VoRTX generates over-smoothed and inaccurate surfaces and has problems yielding the correct geometry for planar surfaces (\eg walls). Due to pixel prediction, SimpleRecon is able to produce more voxels than volumetric methods. However, some meshes generated by SimpleRecon are uneven, caused by depth inconsistency. 
In contrast, our geometry integration mechanism (the 4th column) shows an improvement in reconstruction quality, \ie recovering more surfaces (\eg textureless regions), yielding smoother and more accurate meshes, and providing proper geometry relationships (\eg perpendicular connections between adjacent walls). More qualitative results can be found in the supplementary material.

\noindent \textbf{7-Scenes \cite{shotton2013scene} and TUM RGB-D \cite{sturm12iros}.} Table \ref{tab:generalization} shows the results on 7-Scenes and TUM RGB-D datasets. Although no fine-tuning is applied to these two datasets, the method using our geometry integration mechanism demonstrates an improvement in performance. On 7-Scenes, the F-score of NeuralRecon + ours is better than the other methods, For instance, there are improvements of 7.7\% and 5.3\% when compared to NeuralRecon and VoRTX, respectively. In the case of TUM RGB-D dataset, within the VoRTX framework, incorporating our method results in a 1.6\% increase in F-score and a 1.4\% increase in recall. Qualitative comparisons on the 7-Scenes and TUM RGB-D datasets are provided in the supplementary material.

\noindent \textbf{Efficiency.} The average running time during forward propagation is shown in Table \ref{tab: speed}. Depth-based methods focus on a single key-frame, while volumetric methods run on several key-frames at the same time. Thus, for fairness, only volumetric methods are compared in Table \ref{tab: speed}.

Like \cite{sun2021neuralrecon}, the reconstruction time of a local fragment is divided by the number of keyframes. The methods in Table \ref{tab: speed} are tested on an NVIDIA RTX A6000 GPU and use the same number of key-frames, \ie 9. It can be derived that our geometry integration mechanism increases the reconstruction performance at the cost of speed. However, the inference costs are comparable. Additionally, NeuralRecon + ours is faster than VoRTX. 

\subsection{Ablation study}
In this section, based on NeuralRecon, an ablation study is conducted to assess the effectiveness of our geometry-guided feature learning, geometry-guided adaptive feature fusion, and consistent 3D normal loss on ScanNet.

\begin{table}[t]
\small
\setlength\tabcolsep{2.5pt}
\centering
\begin{tabular}{c|l|ccc}
\hline
  &  & Prec $\uparrow$& Recall $\uparrow$& F-score$\uparrow$ \\ \hline
a &{NeuralRecon}  & 0.683   &  0.478   &  0.560       \\ \hline
b &{+ Trans.}  & 0.678   & 0.488    &0.566         \\ 
c &{+ Trans. + norm.}  & 0.691   & 0.513    & 0.587        \\ 
d &{+ Trans. + norm. + view. (same)}   & {0.686}  & 0.520 &   0.590  \\ 
e &{+ Trans. + norm. + view.}   & \textbf{0.709}  & {0.521} &   {0.598}  \\ 
f &{+ Trans. + norm. + view. + depth} &{0.701}    & \textbf{0.530}  &  \textbf{0.602} \\ \hline
g &{+ geo. (SimpleRecon)} &{0.678}    & {0.496}  &  {0.571} \\ \hline
\end{tabular}
\caption{Ablation study for geometry-guided feature learning. \emph{Trans.} and \emph{norm.} denote the Transformer and projected surface normals. \emph{view.} is the viewing direction and viewing angle. \emph{geo.(SimpleRecon)} refers to the geometry used in SimpleRecon.
}
\label{tab: feat learn}
\end{table}

\begin{table}[t]
\small
\setlength\tabcolsep{2pt}
\centering
\begin{tabular}{c|l|ccc}
\hline
 & & Prec$\uparrow$ & Recall$\uparrow$ & F-score$\uparrow$ \\ \hline
 a&NeuralRecon + G2FL &0.701 &  0.530  & 0.602   \\ \hline
b& + weight $\mu / \sigma$ &  0.703     &   0.534  &     0.605    \\ \hline
c&+ weight $\mu / \sigma$ + rp $\mu / \sigma$ &  0.704     & 0.541    &   0.610      \\ \hline
d& + weight $\mu / \sigma$ + rp $\mu / \sigma$ + proj. tsdf.& 0.707      & 0.539    & 0.609        \\ \hline
 e&+ weight $\mu / \sigma$ + rp $\mu / \sigma$ + vis. & 0.711  &0.541    & 0.612        \\ \hline
 f&+ weight $\mu / \sigma$ + rp $\mu / \sigma$ + proj. occ.& \textbf{0.713} &   \textbf{0.542}  &\textbf{0.614}      \\ \hline
\end{tabular}
\caption{Ablation study for geometry-guided adaptive feature fusion. \emph{weight $\mu$ / $\sigma$} and \emph{rp $\mu$ / $\sigma$} are the mean and standard deviation of attention weight and relative pose distance. \emph{proj. tsdf}, \emph{vis.}, \emph{proj. occ.} are projective TSDF, visibility, projective occupancy.
}
\label{tab: feat fusion}
\end{table}

\noindent \textbf{Geometry-guided feature learning.} Table \ref{tab: feat learn} provides the ablation study for geometry-guided feature learning. In rows \emph{a} and \emph{b}, it's evident that the Transformer blocks enhance both recall and F-score. Moving to row \emph{c}, the incorporation of projected surface normals leads to a 2.1\% increase in F-score. A comparison between rows \emph{c} and \emph{e} highlights the influence of viewing angle and direction, contributing to heightened precision, recall, and F-score. Furthermore, row \emph{d} presents outcomes from combining normal and viewing priors within the same Transformer. In contrast to row \emph{e} where priors are distributed across different modules, row \emph{d} indicates an insufficient exploration of geometric information. Projected depth in row \emph{f} results in a 0.9\% improvement in recall and a 0.4\% improvement in F-score. Row \emph{g} shows the results with geometry used in SimpleRecon, which is worse than ours in row \emph{f}. Finally, compared to NeuralRecon in row \emph{a}, our geometry-guided feature learning in row \emph{f} improves the performance, with an increase in recall by 5.2\% and F-score by 4.2\%. This is attributed to not only the Transformer blocks, but also the geometric priors.

\noindent \textbf{Geometry-guided adaptive feature fusion.} Ablation experiments for geometry-guided adaptive feature fusion are presented in Table \ref{tab: feat fusion}, which are built on Table \ref{tab: feat learn}{{f}}, \ie, $NeuralRecon + G2FL$. In Table \ref{tab: feat fusion}, rows \emph{b}-\emph{f} with adaptively learned weights all outperform $NeuralRecon + G2FL$. Although the Transformer in G2FL can learn attention for multiple views, only using the attention weight $\textbf{A}$ as a weight guidance in row \emph{b} provides 0.3\% improvement in F-score. Furthermore, the relative pose distance in row \emph{c} is able to increase the F-score by another 0.5\%. The \emph{d}-\emph{f} rows explore different representations (\ie  projective TSDF, visibility, and projective occupancy) as approximations to occlusion. In row \emph{d}, F-score decreases slightly. The reason is that the prediction of projective TSDF is a regression task, and the network has difficulty optimizing it. Projective occupancy reaches a better performance than other approximations, not only because it can be used to measure the occlusion but also because the representation is close to the reconstruction task. Compared to $NeuralRecon + G2FL$, the performance of our geometry-guided feature fusion in the last row is increased by 1.2\% in precision, recall, and F-score, which shows the effectiveness of our fusion module. 

\noindent \textbf{Consistent 3D normal loss.} The results in Table \ref{tab: normal loss} validate the effectiveness of our consistent 3D normal loss, which is based on Table \ref{tab: feat fusion}{f}, \ie $NeuralRecon + G2FL + G2AFF$. Experiments of rows \emph{b-d} are conducted to show the importance of consistency weighting and boundary masking. Row \emph{b} shows the results for 3D normal loss without consistency weights and boundary masks, which is worse than row \emph{a}. This means that the normal loss should not be applied to all voxels. Compared to rows \emph{b} and \emph{d}, rows \emph{c} and \emph{f} present an increase in F-score. This indicates that the boundary mask is useful in our 3D normal loss. The recall and F-score in row \emph{d} are 0.5\% and 0.2\% higher than in row \emph{b}. The same trends can also be observed for rows \emph{c} and \emph{f}, which shows that consistent weights play an important role in our 3D normal loss. Row \emph{e} replaces the indicator function in Eq. \ref{eq: consist. weight} by a Gaussian function. The results of our indicator function are better than row \emph{e}. The comparison between rows \emph{f} and \emph{a} demonstrates that our consistent normal loss gives a better reconstruction performance.

\begin{figure*}[t]
  \centering
  \includegraphics[width=0.95\linewidth]{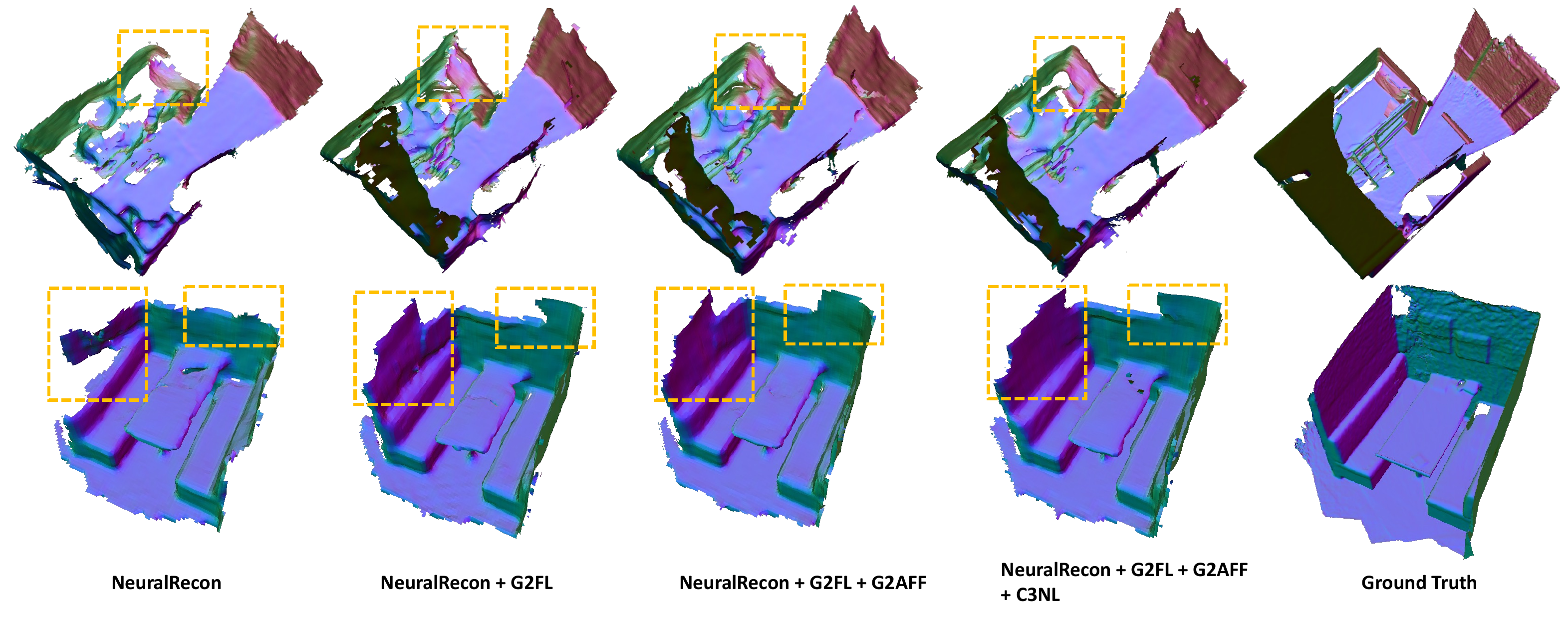}
  \caption{ \textbf{Visualization comparison of the ablation study.} Our proposed geometry-guided feature learning (G2FL), geometry-guided adaptive feature fusion (G2AFF), and consistent 3D normal loss (C3NL) all contribute to an improved reconstruction quality.}
  \label{fig: visualab}
\end{figure*}

\noindent \textbf{Qualitative comparison.} Visualization results of the ablation study are presented in Figure \ref{fig: visualab}. Unlike NeuralRecon, our suggested integration of geometry (refer to columns 2-4) plays a significant role in recovering regions, establishing coherent planes and accurate interrelationships among walls. This underscores the significance of incorporating 3D geometry at various stages.

\begin{table}[t]
\small
\setlength\tabcolsep{2pt}
\centering
\begin{tabular}{c|l|ccc}
\hline
  &  & Prec$\uparrow$ & Recall$\uparrow$ & F-score$\uparrow$ \\ \hline
a &NeuralRecon + G2FL + G2AFF & {0.713} & {0.542}  &{0.614}\\ \hline
b &+ normal loss (w/o weight) & 0.699   &0.542  & 0.609   \\ \hline
c &+ normal loss (w/o consist. weight)  &  0.708 &0.543  & 0.613  \\ \hline
d &+ normal loss (w/o boundary mask)  & 0.698 &0.547   &   0.611 \\ \hline
e &+ normal loss (Gaussian weight)  &  0.705  &  \textbf{0.549}& 0.615    \\ \hline
f & Ours &\textbf{0.722}   & {0.545} & \textbf{0.619}   \\ \hline
\end{tabular}
\caption{Ablation study for consistent 3D normal loss.
}
\label{tab: normal loss}
\end{table}

\begin{table}[t]
\small
\centering
\setlength\tabcolsep{10pt}
\begin{tabular}{c|c|c}
\hline
  Method &  Time $\downarrow$  & F-score $\uparrow$\\ \hline
  NeuralRecon  &   \textbf{27}   &0.560 \\
  
  {NeuralRecon + ours} &  \underline{35} &\underline{0.619}    \\ \hline
  VoRTX  & 37    &0.598    \\ 
  {VoRTX + Ours}  &  41  & \textbf{0.630}     \\ \hline
\end{tabular}
\caption{Comparison of average running time in milliseconds per keyframe.
}
\label{tab: speed}
\end{table}

\section{Conclusion}
In this paper, a novel geometry integration mechanism is presented to explore 3D geometry in indoor scene reconstruction. The key contribution is to encode geometric priors at three levels, \ie feature learning, feature fusion, and network supervision. Geometry-guided feature learning is proposed to integrate view-dependent geometry into the multi-view visual features, enhancing the reconstruction features. The proposed geometry-guided adaptive feature fusion adopts geometry as guidance to model the weight function for multiple views. A consistent 3D normal loss is designed to add local supervision, considering only planar regions and consistency between 2D and 3D normals. 

Large-scale experiments are conducted on the ScanNet dataset, showing that our method outperforms state-of-the-art methods quantitatively as well as qualitatively. Volumetric methods with our geometry integration mechanism also show good generalization on 7-Scenes and TUM RGB-D.

\noindent \textbf{Acknowledgment.} 
The authors deeply appreciate the helpful suggestions on the manuscript from Haochen Wang, Qi Bi, Yongtuo Liu, and Yunlu Chen. The authors also would like to thank all reviewers for their constructive feedback. 

{\small
\bibliographystyle{ieee_fullname}
\bibliography{main}

\begin{thebibliography}{10}\itemsep=-1pt

\bibitem{bae2021estimating}
Gwangbin Bae, Ignas Budvytis, and Roberto Cipolla.
\newblock Estimating and exploiting the aleatoric uncertainty in surface normal estimation.
\newblock In {\em Proceedings of the IEEE/CVF International Conference on Computer Vision}, pages 13137--13146, 2021.

\bibitem{bozic2021transformerfusion}
Aljaz Bozic, Pablo Palafox, Justus Thies, Angela Dai, and Matthias Nie{\ss}ner.
\newblock Transformerfusion: Monocular rgb scene reconstruction using transformers.
\newblock {\em Advances in Neural Information Processing Systems}, 34, 2021.

\bibitem{chu2021twins}
Xiangxiang Chu, Zhi Tian, Yuqing Wang, Bo Zhang, Haibing Ren, Xiaolin Wei, Huaxia Xia, and Chunhua Shen.
\newblock Twins: Revisiting the design of spatial attention in vision transformers.
\newblock {\em Advances in Neural Information Processing Systems}, 34, 2021.

\bibitem{collins1996space}
Robert~T Collins.
\newblock A space-sweep approach to true multi-image matching.
\newblock In {\em Proceedings of the IEEE/CVF Conference on Computer Vision and Pattern Recognition}, pages 358--363, 1996.

\bibitem{curless1996volumetric}
Brian Curless and Marc Levoy.
\newblock A volumetric method for building complex models from range images.
\newblock In {\em Proceedings of the 23rd Annual Conference on Computer Graphics and Interactive Techniques}, pages 303--312, 1996.

\bibitem{dai2017scannet}
Angela Dai, Angel~X Chang, Manolis Savva, Maciej Halber, Thomas Funkhouser, and Matthias Nie{\ss}ner.
\newblock Scannet: Richly-annotated 3d reconstructions of indoor scenes.
\newblock In {\em Proceedings of the IEEE Conference on Computer Vision and Pattern Recognition}, pages 5828--5839, 2017.

\bibitem{duzceker2021deepvideomvs}
Arda Duzceker, Silvano Galliani, Christoph Vogel, Pablo Speciale, Mihai Dusmanu, and Marc Pollefeys.
\newblock Deepvideomvs: Multi-view stereo on video with recurrent spatio-temporal fusion.
\newblock In {\em Proceedings of the IEEE/CVF Conference on Computer Vision and Pattern Recognition}, pages 15324--15333, 2021.

\bibitem{gallup2007real}
David Gallup, Jan-Michael Frahm, Philippos Mordohai, Qingxiong Yang, and Marc Pollefeys.
\newblock Real-time plane-sweeping stereo with multiple sweeping directions.
\newblock In {\em Proceedings of the IEEE/CVF Conference on Computer Vision and Pattern Recognition}, pages 1--8, 2007.

\bibitem{he2016deep}
Kaiming He, Xiangyu Zhang, Shaoqing Ren, and Jian Sun.
\newblock Deep residual learning for image recognition.
\newblock In {\em Proceedings of the IEEE Conference on Computer Vision and Pattern Recognition}, pages 770--778, 2016.

\bibitem{hou2019multi}
Yuxin Hou, Juho Kannala, and Arno Solin.
\newblock Multi-view stereo by temporal nonparametric fusion.
\newblock In {\em Proceedings of the IEEE/CVF International Conference on Computer Vision}, pages 2651--2660, 2019.

\bibitem{im2019dpsnet}
Sunghoon Im, Hae-Gon Jeon, Stephen Lin, and In~So Kweon.
\newblock Dpsnet: End-to-end deep plane sweep stereo.
\newblock {\em arXiv preprint arXiv:1905.00538}, 2019.

\bibitem{irwin1968isotropic}
FG Irwin et~al.
\newblock An isotropic 3x3 image gradient operator.
\newblock {\em Presentation at Stanford AI Project}, 2014(02), 1968.

\bibitem{liu2019neural}
Chao Liu, Jinwei Gu, Kihwan Kim, Srinivasa~G Narasimhan, and Jan Kautz.
\newblock Neural rgb (r) d sensing: Depth and uncertainty from a video camera.
\newblock In {\em Proceedings of the IEEE/CVF Conference on Computer Vision and Pattern Recognition}, pages 10986--10995, 2019.

\bibitem{long2020occlusion}
Xiaoxiao Long, Lingjie Liu, Christian Theobalt, and Wenping Wang.
\newblock Occlusion-aware depth estimation with adaptive normal constraints.
\newblock In {\em Proceedings of European Conference on Computer Vision}, pages 640--657, 2020.

\bibitem{lorensen1987marching}
William~E Lorensen and Harvey~E Cline.
\newblock Marching cubes: A high resolution 3d surface construction algorithm.
\newblock {\em ACM Siggraph Computer Graphics}, 21(4):163--169, 1987.

\bibitem{mildenhall2020nerf}
Ben Mildenhall, Pratul~P Srinivasan, Matthew Tancik, Jonathan~T Barron, Ravi Ramamoorthi, and Ren Ng.
\newblock Nerf: Representing scenes as neural radiance fields for view synthesis.
\newblock In {\em Proceedings of European Conference on Computer Vision}, pages 405--421, 2020.

\bibitem{murez2020atlas}
Zak Murez, Tarrence Van~As, James Bartolozzi, Ayan Sinha, Vijay Badrinarayanan, and Andrew Rabinovich.
\newblock Atlas: End-to-end 3d scene reconstruction from posed images.
\newblock In {\em Proceedings of European Conference on Computer Vision}, pages 414--431, 2020.

\bibitem{newcombe2011kinectfusion}
Richard~A Newcombe, Shahram Izadi, Otmar Hilliges, David Molyneaux, David Kim, Andrew~J Davison, Pushmeet Kohi, Jamie Shotton, Steve Hodges, and Andrew Fitzgibbon.
\newblock Kinectfusion: Real-time dense surface mapping and tracking.
\newblock In {\em 2011 10th IEEE International Symposium on Mixed and Augmented Reality}, pages 127--136, 2011.

\bibitem{peng2020convolutional}
Songyou Peng, Michael Niemeyer, Lars Mescheder, Marc Pollefeys, and Andreas Geiger.
\newblock Convolutional occupancy networks.
\newblock In {\em Proceedings of European Conference on Computer Vision}, pages 523--540, 2020.

\bibitem{prewitt1970object}
Judith~MS Prewitt et~al.
\newblock Object enhancement and extraction.
\newblock {\em Picture Processing and Psychopictorics}, 10(1):15--19, 1970.

\bibitem{rich20213dvnet}
Alexander Rich, Noah Stier, Pradeep Sen, and Tobias H{\"o}llerer.
\newblock 3dvnet: Multi-view depth prediction and volumetric refinement.
\newblock In {\em 2021 International Conference on 3D Vision}, pages 700--709, 2021.

\bibitem{robert2022learning}
Damien Robert, Bruno Vallet, and Loic Landrieu.
\newblock Learning multi-view aggregation in the wild for large-scale 3d semantic segmentation.
\newblock In {\em Proceedings of the IEEE/CVF Conference on Computer Vision and Pattern Recognition}, pages 5575--5584, 2022.

\bibitem{sayed2022simplerecon}
Mohamed Sayed, John Gibson, Jamie Watson, Victor Prisacariu, Michael Firman, and Cl{\'e}ment Godard.
\newblock Simplerecon: 3d reconstruction without 3d convolutions.
\newblock In {\em Proceedings of European Conference on Computer Vision}, pages 1--19, 2022.

\bibitem{schonberger2016pixelwise}
Johannes~L Sch{\"o}nberger, Enliang Zheng, Jan-Michael Frahm, and Marc Pollefeys.
\newblock Pixelwise view selection for unstructured multi-view stereo.
\newblock In {\em Proceedings of European Conference on Computer Vision}, pages 501--518, 2016.

\bibitem{seeger2004gaussian}
Matthias Seeger.
\newblock Gaussian processes for machine learning.
\newblock {\em International Journal of Neural Systems}, 14(02):69--106, 2004.

\bibitem{shi2015convolutional}
Xingjian Shi, Zhourong Chen, Hao Wang, Dit-Yan Yeung, Wai-Kin Wong, and Wang-chun Woo.
\newblock Convolutional lstm network: A machine learning approach for precipitation nowcasting.
\newblock {\em Advances in Neural Information Processing Systems}, 28, 2015.

\bibitem{shotton2013scene}
Jamie Shotton, Ben Glocker, Christopher Zach, Shahram Izadi, Antonio Criminisi, and Andrew Fitzgibbon.
\newblock Scene coordinate regression forests for camera relocalization in rgb-d images.
\newblock In {\em Proceedings of the IEEE Conference on Computer Vision and Pattern Recognition}, pages 2930--2937, 2013.

\bibitem{stier2021vortx}
Noah Stier, Alexander Rich, Pradeep Sen, and Tobias H{\"o}llerer.
\newblock Vortx: Volumetric 3d reconstruction with transformers for voxelwise view selection and fusion.
\newblock In {\em 2021 International Conference on 3D Vision}, pages 320--330, 2021.

\bibitem{sturm12iros}
J. Sturm, N. Engelhard, F. Endres, W. Burgard, and D. Cremers.
\newblock A benchmark for the evaluation of rgb-d slam systems.
\newblock In {\em Proceedings of the International Conference on Intelligent Robot Systems}, pages 573--580, 2012.

\bibitem{sun2021neuralrecon}
Jiaming Sun, Yiming Xie, Linghao Chen, Xiaowei Zhou, and Hujun Bao.
\newblock Neuralrecon: Real-time coherent 3d reconstruction from monocular video.
\newblock In {\em Proceedings of the IEEE/CVF Conference on Computer Vision and Pattern Recognition}, pages 15598--15607, 2021.

\bibitem{tan2019mnasnet}
Mingxing Tan, Bo Chen, Ruoming Pang, Vijay Vasudevan, Mark Sandler, Andrew Howard, and Quoc~V Le.
\newblock Mnasnet: Platform-aware neural architecture search for mobile.
\newblock In {\em Proceedings of the IEEE/CVF Conference on Computer Vision and Pattern Recognition}, pages 2820--2828, 2019.

\bibitem{tang2020searching}
Haotian Tang, Zhijian Liu, Shengyu Zhao, Yujun Lin, Ji Lin, Hanrui Wang, and Song Han.
\newblock Searching efficient 3d architectures with sparse point-voxel convolution.
\newblock In {\em Proceedings of European Conference on Computer Vision}, pages 685--702, 2020.

\bibitem{tang2022contrastive}
Liyao Tang, Yibing Zhan, Zhe Chen, Baosheng Yu, and Dacheng Tao.
\newblock Contrastive boundary learning for point cloud segmentation.
\newblock In {\em Proceedings of the IEEE/CVF Conference on Computer Vision and Pattern Recognition}, pages 8489--8499, 2022.

\bibitem{teed2018deepv2d}
Zachary Teed and Jia Deng.
\newblock Deepv2d: Video to depth with differentiable structure from motion.
\newblock {\em arXiv preprint arXiv:1812.04605}, 2018.

\bibitem{vaswani2017attention}
Ashish Vaswani, Noam Shazeer, Niki Parmar, Jakob Uszkoreit, Llion Jones, Aidan~N Gomez, {\L}ukasz Kaiser, and Illia Polosukhin.
\newblock Attention is all you need.
\newblock {\em Advances in Neural Information Processing Systems}, 30, 2017.

\bibitem{vu2022softgroup}
Thang Vu, Kookhoi Kim, Tung~M Luu, Thanh Nguyen, and Chang~D Yoo.
\newblock Softgroup for 3d instance segmentation on point clouds.
\newblock In {\em Proceedings of the IEEE/CVF Conference on Computer Vision and Pattern Recognition}, pages 2708--2717, 2022.

\bibitem{wang2022cagroup3d}
Haiyang Wang, Lihe Ding, Shaocong Dong, Shaoshuai Shi, Aoxue Li, Jianan Li, Zhenguo Li, and Liwei Wang.
\newblock Cagroup3d: Class-aware grouping for 3d object detection on point clouds.
\newblock {\em arXiv preprint arXiv:2210.04264}, 2022.

\bibitem{wang2018mvdepthnet}
Kaixuan Wang and Shaojie Shen.
\newblock Mvdepthnet: Real-time multiview depth estimation neural network.
\newblock In {\em 2018 International Conference on 3D Vision}, pages 248--257, 2018.

\bibitem{wang2021pyramid}
Wenhai Wang, Enze Xie, Xiang Li, Deng-Ping Fan, Kaitao Song, Ding Liang, Tong Lu, Ping Luo, and Ling Shao.
\newblock Pyramid vision transformer: A versatile backbone for dense prediction without convolutions.
\newblock In {\em Proceedings of the IEEE/CVF International Conference on Computer Vision}, pages 568--578, 2021.

\bibitem{wang2022multimodal}
Yikai Wang, Xinghao Chen, Lele Cao, Wenbing Huang, Fuchun Sun, and Yunhe Wang.
\newblock Multimodal token fusion for vision transformers.
\newblock In {\em Proceedings of the IEEE/CVF Conference on Computer Vision and Pattern Recognition}, pages 12186--12195, 2022.

\bibitem{yang2020robust}
Bo Yang, Sen Wang, Andrew Markham, and Niki Trigoni.
\newblock Robust attentional aggregation of deep feature sets for multi-view 3d reconstruction.
\newblock {\em International Journal of Computer Vision}, 128(1):53--73, 2020.

\bibitem{yao2018mvsnet}
Yao Yao, Zixin Luo, Shiwei Li, Tian Fang, and Long Quan.
\newblock Mvsnet: Depth inference for unstructured multi-view stereo.
\newblock In {\em Proceedings of the European Conference on Computer Vision}, pages 767--783, 2018.

\bibitem{yao2019recurrent}
Yao Yao, Zixin Luo, Shiwei Li, Tianwei Shen, Tian Fang, and Long Quan.
\newblock Recurrent mvsnet for high-resolution multi-view stereo depth inference.
\newblock In {\em Proceedings of the IEEE/CVF Conference on Computer Vision and Pattern Recognition}, pages 5525--5534, 2019.

\bibitem{yi2020pyramid}
Hongwei Yi, Zizhuang Wei, Mingyu Ding, Runze Zhang, Yisong Chen, Guoping Wang, and Yu-Wing Tai.
\newblock Pyramid multi-view stereo net with self-adaptive view aggregation.
\newblock In {\em Proceedings of European Conference on Computer Vision}, pages 766--782, 2020.

\bibitem{zhang2021vidtr}
Yanyi Zhang, Xinyu Li, Chunhui Liu, Bing Shuai, Yi Zhu, Biagio Brattoli, Hao Chen, Ivan Marsic, and Joseph Tighe.
\newblock Vidtr: Video transformer without convolutions.
\newblock In {\em Proceedings of the IEEE/CVF International Conference on Computer Vision}, pages 13577--13587, 2021.

\end{thebibliography}
}

\clearpage
\vspace{5mm}
{\Large\centering \textbf{Supplementary Material}} 
\vspace{5mm}
\appendix

\renewcommand*{\thefigure}{S\arabic{figure}}
\renewcommand*{\thetable}{S\arabic{table}}
\renewcommand*{\theequation}{S-\arabic{equation}}
\setcounter{table}{0}
\setcounter{figure}{0}
\setcounter{equation}{0}

This supplementary document provides additional details and experimental results of our geometry integration mechanism. Section \ref{sec: metrics} presents the definitions of 3D evaluation metrics. Section \ref{sec: geometry} details the computation of geometric priors. Details of network implementation are given in Section \ref{sec: network}. Section \ref{sec: results} shows additional geometry measures, ablation study, analysis, and visualizations. 

\section{Evaluation metrics}
\label{sec: metrics}
Table \ref{tab: metrics} presents the definitions of 3D evaluation metrics in Atlas \cite{murez2020atlas}, \ie accuracy (acc), completeness (comp), precision (prec), recall, and F-score. For accuracy and completeness, lower is better. Conversely, for the remaining metrics, higher values correspond to better performance.

\section{Geometric priors}
\label{sec: geometry}
Geometric priors used in our method are introduced in this section.

\noindent {\textbf{Viewing direction $\textbf{v}_i$:}} The normalized unit direction from the camera origin to the 3D voxel. In our method, it is encoded similarly to NeRF \cite{mildenhall2020nerf}, \ie,
\begin{equation}
\begin{aligned}
    \gamma(\textbf{v}_i)=&(\text{sin}(2^0 \pi \textbf{v}_i), \text{cos}(2^0 \pi \textbf{v}_i), \dots,\\
    &\text{sin}(2^{L-1} \pi \textbf{v}_i),\text{cos}(2^{L-1} \pi \textbf{v}_i))
\end{aligned}
\label{eq: nerf}
\end{equation}
where $\gamma(\cdot)$ is applied to each dimension of $\textbf{v}_i$, and $L=4$ in our experiments.

\noindent {\textbf{Projected normal $\textbf{n}_i$:}} The vector mapped from the 2D normal according to perspective projection.

\noindent {\textbf{Viewing angle $\boldsymbol{\theta}_i$:}} The absolute value of the cosine similarity between the projected normal and the viewing direction. 

\noindent {\textbf{Projected depth $\textbf{z}_i$:}} The perpendicular distance from the 3D voxel to the camera center. Our method divides the distance by the maximum depth $D_{max}=3\text{m}$. The projected depth is also encoded using Eq. \ref{eq: nerf} with $L=4$.

\noindent {\textbf{Relative pose distance:}} The pose distance between two cameras. The overall pose distance between camera $j$ and camera $k$ is calculated by 
\begin{equation}
\begin{aligned}
rp_{jk}
&=\sqrt{||\textbf{t}_{jk}||^2+\frac{2}{3}\text{tr}(\mathbb{I}-\textbf{R}_{jk})} \\
&=\sqrt{rp(\textbf{t}_{jk})^2 + rp(\textbf{R}_{jk})^2}
\end{aligned}
\end{equation}
where $\textbf{t}_{jk}$ is the relative translation matrix, $\textbf{R}_{jk}$ is the relative rotation matrix. $rp(\textbf{t}_{jk})$ denotes the pose translation distance. $rp(\textbf{R}_{jk})$ denotes the pose rotation distance. \emph{tr} is the matrix trace operator. In our proposed geometry-guided adaptive feature fusion, $rp(\textbf{R}_{jk})$, $rp(\textbf{t}_{jk})$, and $rp_{jk}$ are all used as priors to learn the weight function.

\noindent {\textbf{Projective occupancy:}} In a camera coordinate frame, the projective TSDF $S_p(\textbf{p})$ of a voxel is the truncated signed distance between the voxel $\textbf{p}$ and the nearest surface. Projective occupancy $O(\textbf{p})$ and visibility $V(\textbf{p})$ are functions of projective TSDF, which can be written by
\begin{equation}
\begin{aligned}
&O(\textbf{p})=[|S_p(\textbf{p})|< t] \equiv 
\begin{cases}
        1,& |S_p(\textbf{p})|< t \\
        0, & |S_p(\textbf{p})|\ge t
    \end{cases} \\
&V(\textbf{p})=[S_p(\textbf{p})\ge 0] \equiv 
\begin{cases}
        1,& S_p(\textbf{p})\ge0 \\
        0, & S_p(\textbf{p})< 0
    \end{cases} \\
\end{aligned}
\end{equation}
where $t$ is the truncation distance. 

\begin{table}[t]
\centering
\setlength\tabcolsep{10pt}
\begin{tabular}{ll}
\hline
  \multicolumn{2}{c}{\textbf{3D Metrics}} \\ \hline
  Acc&    $\text{mean}_{p \in P}(\text{min}_{p^* \in {P^*}}||p-p^*||)$ \\ 
  Comp&   $\text{mean}_{p^* \in P^*}(\text{min}_{p \in {P}}||p-p^*||)$    \\ 
  Prec& $\text{mean}_{p \in P}(\text{min}_{p^* \in {P^*}}||p-p^*||<.05)$  \\ 
  Recall& $\text{mean}_{p^* \in P^*}(\text{min}_{p \in {P}}||p-p^*||<.05)$ \\
  F-score & $\frac{2\times \text{Prec} \times \text{Recall}}{\text{Prec} + \text{Recall}}$ \\ \hline
\end{tabular}
\caption{Definitions of 3D metrics. $p$ and $p^*$ are the predicted and ground truth point clouds.
}
\label{tab: metrics}
\end{table}

\section{Implementation details}
\label{sec: network}
In our geometry-guided feature learning, the MLP is composed of 2 linear layers and 2 ReLUs, with channel sizes [37, 32, 1]. The 37 input channels consist of $3\times8$ encoded viewing directions, $1\times8$ encoded projected depths, 3 viewing directions, 1 projected depth, and 1 viewing angle. The channel size of linear layer in $\mathcal{T}_1$ is [$C_v + 1$, $C_v$]. The linear layer in $\mathcal{T}_2$ has [$C_v + 3$, $C_v$] channels. 

In our geometry-guided adaptive feature fusion, the MLP includes 3 linear layers, with two ReLUs following the first two linear layers. The channel sizes are [81, 32, 32, 9], where the input channel 81 is composed of $2\times 9$ for the mean and standard deviation of the attention matrix, $6\times 9$ for the mean and standard deviation of the relative pose distance, and $1\times 9$ for the occlusion prior. The channel of the linear layer for projective occupancy prediction is [$C_v$, 1].

In the ablation study for our consistent 3D normal loss, the Gaussian function in Table {5e} is defined by $\text{exp}(-\frac{(s_{2d3d}(\textbf{p})-1)^2}{\sigma ^2})$, in which $\sigma^2=0.5$.
\section{Additional results}
\label{sec: results}

\begin{table}[t]
\small
\setlength\tabcolsep{3.5pt}
\centering
\begin{tabular}{l|ccc}
\hline
  & Prec $\uparrow$& Recall $\uparrow$& F-score$\uparrow$ \\ \hline
{NeuralRecon + G2FL (w/o)}  & 0.697   &  0.530   &  0.600       \\ 
{NeuralRecon + G2FL(w)}  & 0.701   & 0.530    &0.602         \\ \hline
\end{tabular}
\caption{Ablation study for the encoding function defined in Eq. \ref{eq: nerf}. \emph{w/o} and \emph{w} mean without and with the encoding function respectively.
}
\label{tab: nerf}
\end{table}

\begin{table}[t]
\setlength\tabcolsep{1.2pt}
\footnotesize
\centering
\begin{tabular}{c|cccccc}
\hline
 & P$_{11.25}\uparrow$ & R$_{11.25}\uparrow$  & P$_{22.5} \uparrow$& R$_{22.5} \uparrow$&  P$_{30}\uparrow$ &R$_{30}\uparrow$\\ \hline
{NeuralRecon}& 0.501&  0.418 & 0.697   &  0.608   &  0.764&  0.679    \\ 
{NeuralRecon + ours} & \textbf{0.581} & \underline{0.504} & \textbf{0.753} & \underline{0.680}    &\textbf{0.809}  &  \underline{0.742}     \\ \hline
{VoRTX}  & 0.515 & 0.479   &    0.698 & 0.663  & 0.757 &  0.726    \\ 
{VoRTX + ours}  &\underline{0.552} & \textbf{0.528} & \underline{0.719}  & \textbf{0.701} &\underline{0.777}  &\textbf{0.761} \\ \hline
\end{tabular}
\caption{3D normal evaluation. 
}
\label{tab: normal}

\end{table}

\begin{figure*}[t]
  \centering

  \includegraphics[width=1.0\linewidth]{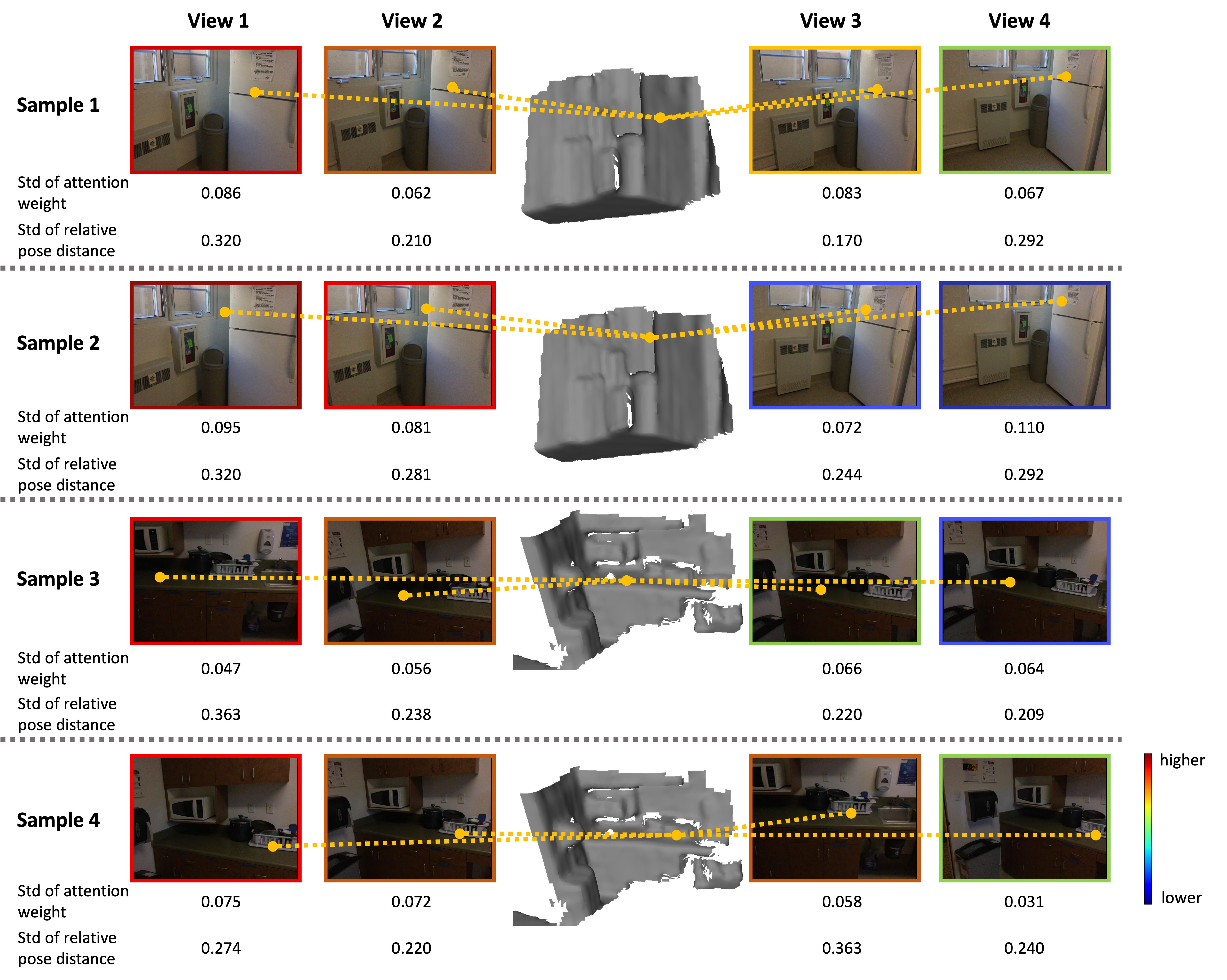}
  \caption{\textbf{Visualization for weight learning in our geometry-guided adaptive feature fusion.}  The view weight of each sample decreases progressively from \emph{View 1} to \emph{View 4}, with the color of the image box as a clearer indication of the weight. The standard deviation (std) of attention weight and relative pose distance is also given. The yellow dots represent the sampled 3D voxels and their corresponding 2D pixels. 
  }
  \label{fig: weight}
\end{figure*}

\noindent \textbf{3D normal estimation.} To further demonstrate that our approach can reconstruct more accurate geometry, this work also provides fine-grained geometry measures, \ie precision $P_\tau$ and recall $R_\tau$ of the 3D normal, see Eq. \ref{eq: p_n} and Eq. \ref{eq: r_n}. The 3D vertex normals are generated from meshes, and evaluated following the metrics in \cite{bae2021estimating}. The experimental results in Table \ref{tab: normal} demonstrate that our proposed geometry integration mechanism enhances the normal performance, thereby contributing to the reconstruction of more precise and accurate geometry.
\begin{equation}
\begin{aligned}
    P_{\tau} &= \text{mean}_{p \in P}(\text{angle}(p, p^*)<\tau), \\
    p^* &= \text{min}_{p^* \in {P^*}}||p-p^*|| 
\end{aligned}
\label{eq: p_n}
\end{equation}

\begin{equation}
\begin{aligned}
    R_{\tau} &= \text{mean}_{p^* \in P^*}(\text{angle}(p, p^*)<\tau), \\
    p &= \text{min}_{p \in {P}}||p-p^*|| 
\end{aligned}
\label{eq: r_n}
\end{equation}
where $p$ and $p^*$ are the predicted and ground truth points. $angle$ is computed between ground truth and prediction. $\tau$ is angle threshold, $\tau \in  \{11.25^{\circ},22.5^{\circ},30^{\circ}\}$.

\noindent \textbf{Additional ablation study.} Table \ref{tab: nerf} validates the effectiveness of the NeRF-like encoding function in our geometry-guided feature learning. As can be seen, with the encoding defined in Eq. \ref{eq: nerf}, precision and F-score increase by 0.4\% and 0.2\% respectively.

\noindent \textbf{Analysis for G2AFF.} The visualization of weight learning in our geometry-guided adaptive feature fusion is presented in Figure \ref{fig: weight}. In \emph{Sample 1}, the highest weight is assigned to the view with the largest standard deviations of attention weight and relative pose distance. In \emph{Sample 2}, due to occlusion in \emph{View 3} and \emph{View 4}, the voxel weights in these two views are very low. \emph{Sample 3} assigns a higher weight to the view with a larger standard deviation of relative pose distance, while \emph{Sample 4} gives more attention to the view with a larger standard deviation of attention weight. In conclusion, our G2AFF is able to learn appropriate weights from the 3D geometry.

\begin{figure*}[t]
  \centering
  \includegraphics[width=0.8\linewidth]{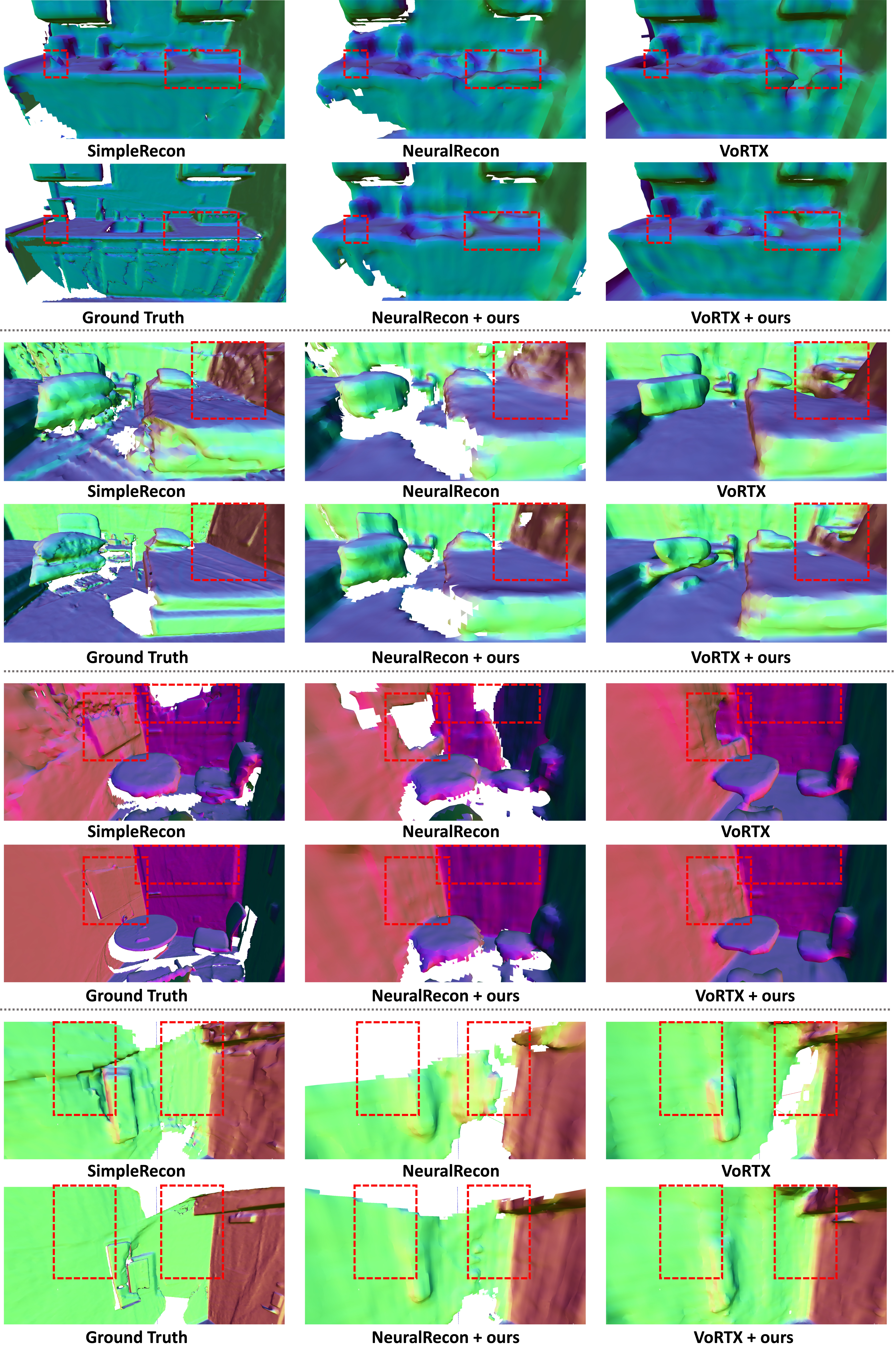}
  \caption{\textbf{Qualitative results on ScanNet.}
  }
  \label{fig: scannet}
\end{figure*}

\noindent \textbf{Additional qualitative results on ScanNet \cite{dai2017scannet}.} More visualizations on ScanNet are shown in Figure \ref{fig: scannet}. Compared to SOTA methods (\ie SimpleRecon \cite{sayed2022simplerecon}, NeuralRecon \cite{sun2021neuralrecon}, and VoRTX \cite{stier2021vortx}), NeuralRecon + ours is able to reconstruct better meshes. Compared to VoRTX, VoRTX + ours can recall more regions and generate flatter meshes, \eg for walls. It can be demonstrated that our geometry integration mechanism is helpful and can be plugged into both online (\eg NeuralRecon) and offline (\eg VoRTX) volumetric methods.

\begin{figure*}[t]
  \centering

  \includegraphics[width=1.0\linewidth]{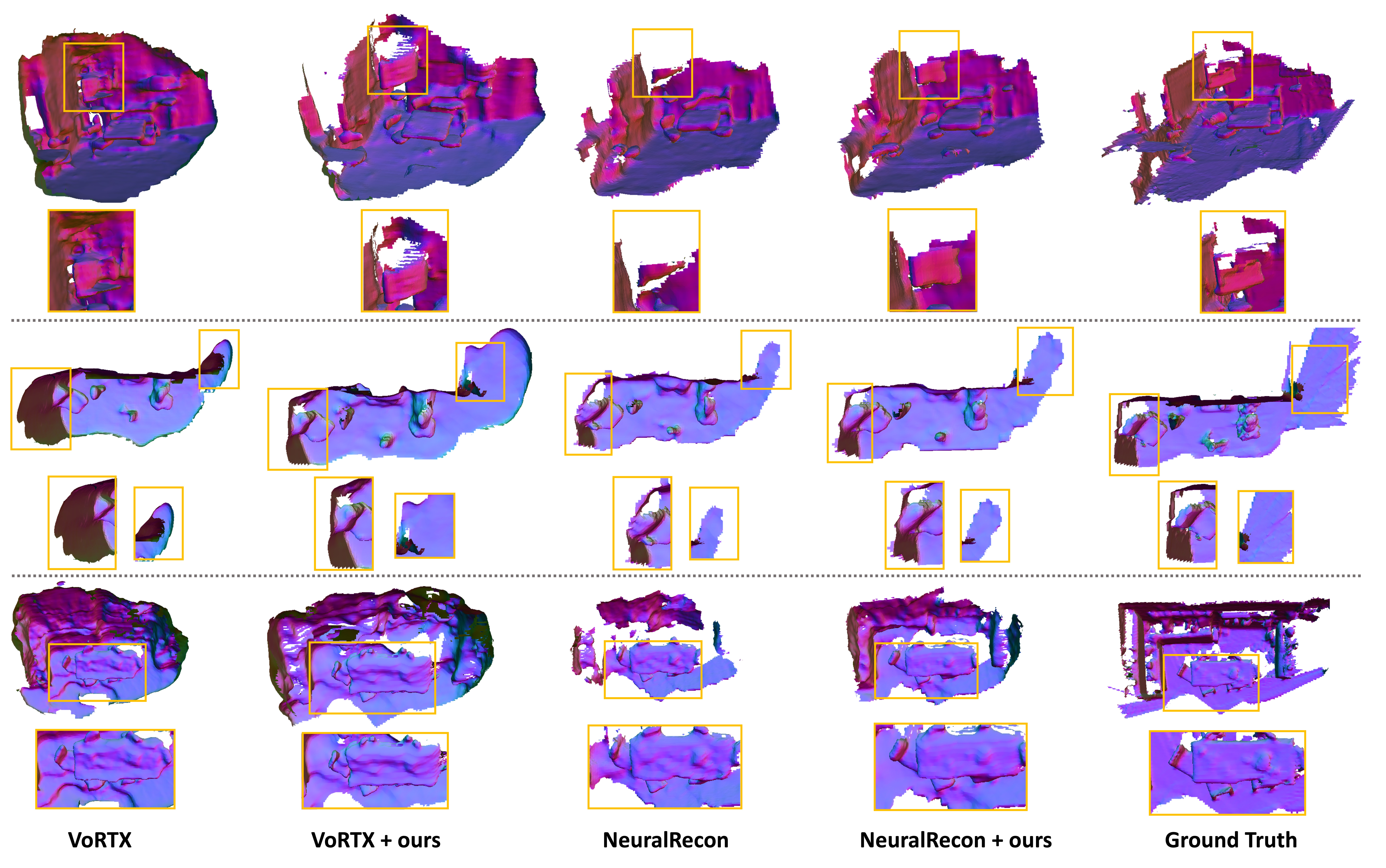}

  \caption{\textbf{Qualitative results on 7-Scenes.}
  }
  \label{fig: 7scenes}
\end{figure*}

\begin{figure*}[t]
  \centering
  \includegraphics[width=0.9\linewidth]{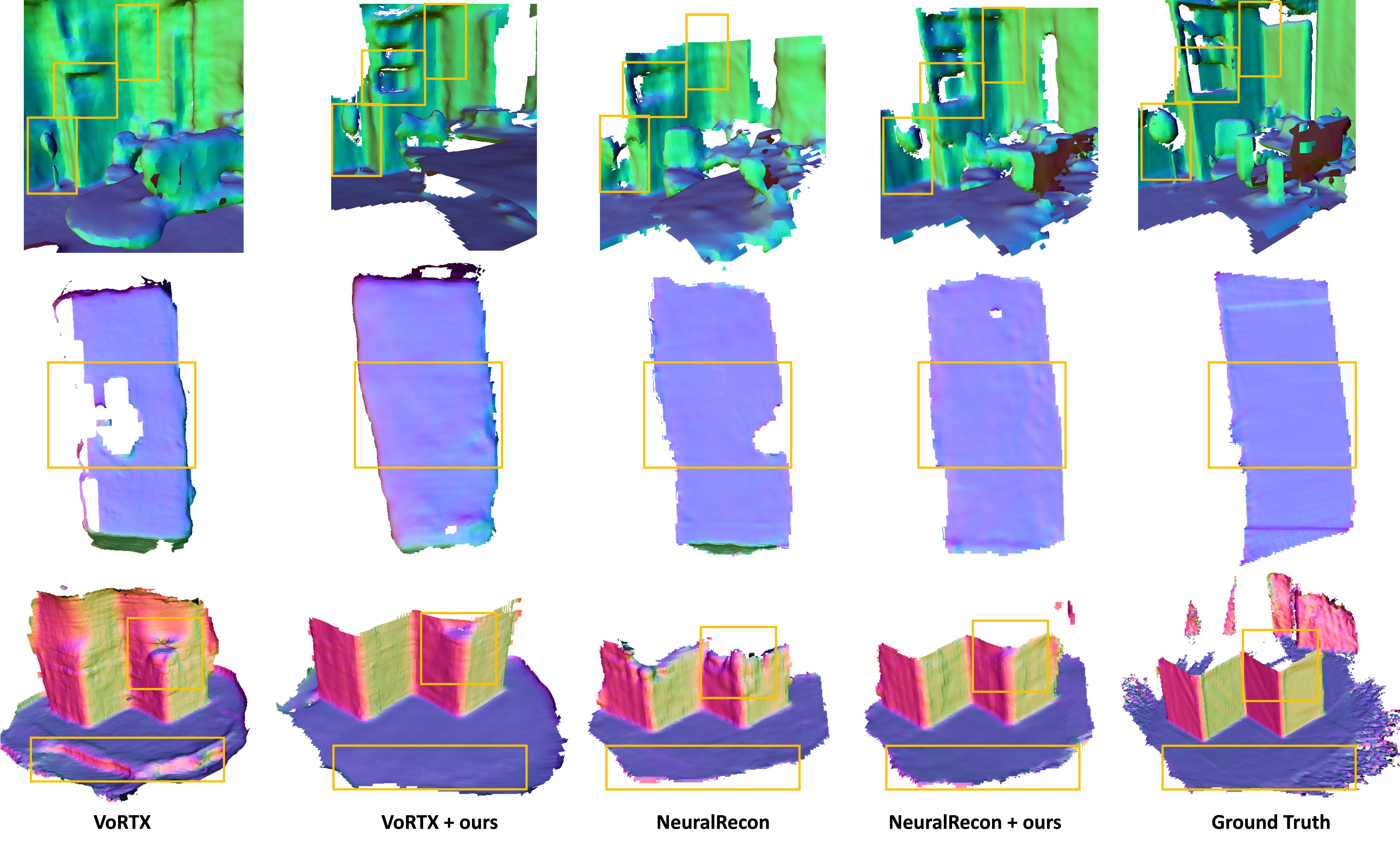}

  \caption{\textbf{Qualitative results on TUM RGB-D.}
  }
  \label{fig: tum}
\end{figure*}

\noindent \textbf{Qualitative results on 7-Scenes \cite{shotton2013scene} and TUM RGB-D \cite{sturm12iros}.} Qualitative results on 7-Scenes and TUM RGB-D datasets are given in Figure \ref{fig: 7scenes} and Figure \ref{fig: tum}. Compared to NeuralRecon and VoRTX, our proposed geometry integration mechanism can recall more meshes and reconstruct flatter planes, \eg for the wall and floor.

\end{document}